\documentclass[letterpaper, 10 pt, journal, twoside]{IEEEtran}
\usepackage{amsmath,amsfonts}
\usepackage{array}
\usepackage[caption=false]{subfig}
\usepackage{textcomp}
\usepackage{stfloats}
\usepackage{url}
\usepackage{verbatim}
\usepackage{graphicx}
\usepackage{cite}
\usepackage{hyperref}
\usepackage{amsmath}
\usepackage{multirow}
\usepackage[ruled,linesnumbered]{algorithm2e}
\usepackage{booktabs} 
\usepackage{array}

\begin{document}

\title{A Spatiotemporal Hand-Eye Calibration for Trajectory Alignment in \\Visual(-Inertial) Odometry Evaluation}

\author{Zichao Shu$^{1}$, Lijun Li$^{1}$, Rui Wang$^{2}$ and Zetao Chen$^{1}$
\thanks{Manuscript received: December 12, 2023; Revised: March 7, 2024; Accepted: March 30, 2024. }
\thanks{This paper was recommended for publication by Editor Sven Behnke upon evaluation of the Associate Editor and Reviewers’ comments. }
\thanks{$^{1}$Zichao Shu, Lijun Li and Zetao Chen are with the Advanced Display and Sensing Research Center, Yongjiang Laboratory, China. \tt\footnotesize\{zichao-shu; lijun-li; zetao-chen\}@ylab.ac.cn}
\thanks{$^{2}$Rui Wang is with the Mixed Reality \& AI Lab-Z\"urich, Microsoft, Switzerland. \tt\footnotesize\{wangr@microsoft.com\}}
\thanks{Digital Object Identifier (DOI): see top of this page. }
}

\markboth{IEEE Robotics and Automation Letters. Preprint Version. Accepted March, 2024}
{Shu \MakeLowercase{\textit{et al.}}: Spatiotemporal Hand-Eye Calibration in Visual(-Inertial) Odometry Evaluation}

\maketitle

\begin{abstract}
A common prerequisite for evaluating a visual(-inertial) odometry (VO/VIO) algorithm is to align the timestamps and the reference frame of its estimated trajectory with a reference ground-truth derived from a system of superior precision, such as a motion capture system. The trajectory-based alignment, typically modeled as a classic hand-eye calibration, significantly influences the accuracy of evaluation metrics. However, traditional calibration methods are susceptible to the quality of the input poses. Few studies have taken this into account when evaluating VO/VIO trajectories that usually suffer from noise and drift.  To fill this gap, we propose a novel spatiotemporal hand-eye calibration algorithm that fully leverages multiple constraints from screw theory for enhanced accuracy and robustness. Experimental results show that our algorithm has better performance and is less noise-prone than state-of-the-art methods. 
\end{abstract}


\section{INTRODUCTION} \label{SEC: INTRODUCTION}

\IEEEPARstart{V}{isual(-inertial)} odometry (VO/VIO) is known to provide state estimation of motion devices and has a wide range of application domains, such as robotics, extended reality, and autonomous driving. The performance evaluation is a fundamental task in VO/VIO research and application, where metrics are typically quantified by evaluating the estimated trajectory from VO/VIO with respect to the ground-truth. Commonly, the ground-truth trajectory can be obtained by tracking the motion device simultaneously with a system of superior precision, e.g., using a motion capture (MoCap) system, laser tracker, etc \cite{burri2016euroc, schuberttum, jinyu2019survey}. There are two main problems when comparing the estimated trajectory against the ground-truth: the trajectory pair is usually on different clock domains (thus with non-corresponding timestamps) and expressed in different global and local reference frames. While well-known methods such as the Umeyama algorithm \cite{umeyama1991least} can align the global frames, the spatiotemporal alignment, which calculates the offsets of timestamps and local frames of the trajectory pair, still needs meticulous handling. 

The spatiotemporal alignment problem above can be modeled as a classic hand-eye calibration problem: given the local frames of the ground-truth and estimated trajectory as the hand and the eye respectively, calculate the timestamp offset and estimate the homogeneous transformation between them. While the essence of hand-eye calibration problem has been well addressed in numerous studies \cite{jiang2022overview, pedrosa2021general, koide2019general, ali2019methods, sarabandi2022hand, furrer2018evaluation}, the accuracy and robustness may still be compromised in practical applications. The error introduced in this step will affect the transformation of the ground-truth trajectory, thereby exerting a substantial influence on the subsequent evaluation metrics. 

\subsection{Motivation}
In this work, we consider the scenario in which only the trajectory information is available. This is common among commercial consumer devices, such as extended reality headsets or home robots, where the original raw sensor data used to derive the device trajectories are not accessible by users. Existing hand-eye calibration algorithms can be categorized into two distinct approaches: tightly-coupled and loosely-coupled. The former typically joints raw data from the sensors such as images with information of the calibration boards \cite{pedrosa2021general, koide2019general, ali2019methods} or IMU measurements \cite{burri2016euroc, schuberttum}, and optimizes the result in a maximum likelihood estimation (MLE) framework. The latter, on the other hand, directly calculates the offset between the hand and the eye based on their independently estimated poses \cite{sarabandi2022hand, furrer2018evaluation}. While tightly-coupled approaches can theoretically achieve higher accuracy and are used in well-known benchmarks such as EuRoC \cite{burri2016euroc} and TUM-VI \cite{schuberttum}, they are not applicable in cases where only the trajectory information is available. Loosely-coupled approaches can perform calibration in the pose-only condition, but due to the ubiquitous noise and accumulated error in the VO/VIO estimation, the accuracy and robustness of existing methods are generally insufficient. 

\subsection{Contribution}
\begin{figure}[t]
    \centering
    \includegraphics[width=0.95\linewidth]{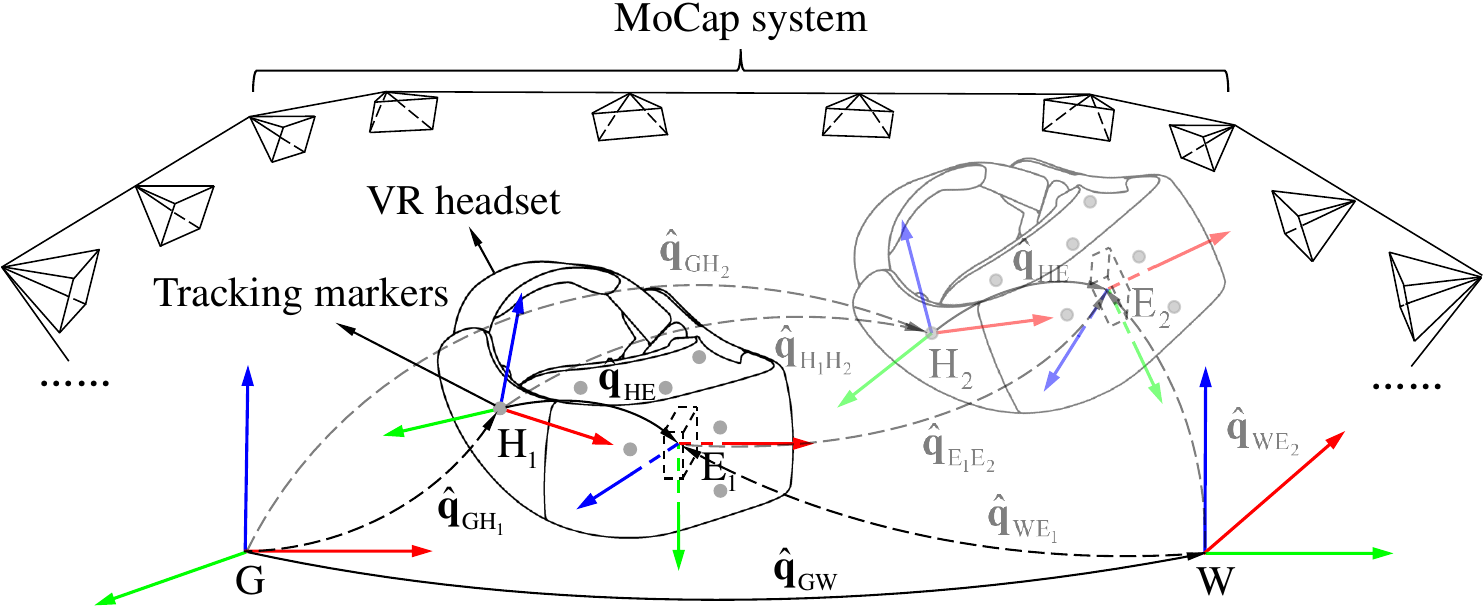}
    \caption{Our spatiotemporal hand-eye calibration platform and the convention of the reference frames. The global frame for MoCap trajectory is denoted as G, and the local frame is referenced to a specific tracking marker indicated as H. The global frame for VIO trajectory is denoted as W, and the local frame E coincides with the IMU body frame. During this process, dashed lines represent transformations that change over time, while solid lines indicate static offsets.}
    \label{FIG: Our hand-eye calibration platform}
\end{figure}

In this letter, we propose a novel loosely-coupled spatiotemporal hand-eye calibration method tailored for VO/VIO evaluation. This method demonstrates robustness against noise and accumulated error in the input trajectories. For time alignment, we improve the correlation analysis of the screw invariant and obtain synchronized trajectories. For spatial calibration, we construct linear equations using local relative poses based on rotational constraint to fully utilize the motion information, rather than the naive global or inter-frame strategies \cite{sarabandi2022hand, ali2019methods, furrer2018evaluation}. Additionally, we introduce a well-designed robust kernel based on the screw theory to stabilize the linear solution. These operations are iteratively completed within the random sample consensus (RANSAC) framework to recover inlier data. Finally, we design a nonlinear optimization tool to jointly refine the time offset and the linear extrinsic solution. To validate the effectiveness of our algorithm, we conduct experiments on public and simulated datasets, as well as our own datasets collected by a virtual reality (VR) headset with VIO capability and a MoCap system (see in Fig. \ref{FIG: Our hand-eye calibration platform}).

The rest of this letter is organized as follows. Section \ref{SEC: RELATED WORK} reviews related work on spatiotemporal hand-eye calibration. The new method is described in Section \ref{SEC: METHODOLOGY} and its performance is evaluated in Section \ref{SEC: EXPERIMENTAL RESULTS}. Section \ref{SEC: CONCLUSIONS} concludes the letter.

\section{RELATED WORK} \label{SEC: RELATED WORK}

Spatiotemporal hand-eye calibration based on different strategies is a widely studied area. In our review, we briefly discuss the related work that shares the same strategy as the proposed method, i.e., the loosely-coupled methods, and motivate the design adapted in our work. 

\subsection{Spatial Hand-Eye Calibration}
In our application scenario, the loosely-coupled hand-eye calibration can be formulated as $\mathbf{AX}$ = $\mathbf{XB}$ \cite{shiu1987calibration}, where $\mathbf{A}$ and $\mathbf{B}$ are the hand and the eye poses between two frames respectively, and $\mathbf{X}$ is the unknown homogeneous transformation between the hand and the eye. The solution for $\mathbf{X}$ can be categorized into two approaches: either separately or simultaneously solving the rotation and translation parts for the transformation.

One of the earliest separated approaches was presented by Shiu and Ahmad \cite{shiu1987calibration}. They represented rotation using angle-axis and proposed a closed-form solution for the $\mathit{\mathbf{AX}}$ = $\mathit{\mathbf{XB}}$ formulation. Tsai and Lenz \cite{tsai1988real} proposed a similar but simplified method to improve computational efficiency, which has been widely used to this day (e.g. in OpenCV). Later methods focused on utilizing various parameterizations of the rotation, such as angle-axis \cite{wang1992extrinsic}, quaternion \cite{chou1991finding}, Lie algebra \cite{park1994robot}, and Kronecker product \cite{liang2008hand}, to achieve more efficient solution. Although separated methods are computationally efficient, they are error-prone due to the independence assumption between the rotation and translation, which are actually nonlinearly coupled \cite{nguyen2018covariance}. 

In contrast, the simultaneous methods calibrate the rotation and translation parts jointly. Representative methods include screw motion \cite{chen1991screw}, Kronecker product \cite{andreff1999line}, dual quaternion \cite{daniilidis1999hand} and dual tensor \cite{condurache2016orthogonal}, which use alternative analytical parameterizations to express the complete homogeneous transformation. Additionally, there are also algorithms based on numerical optimization \cite{strobl2006optimal, heller2014hand, zhao2019simultaneous}. While the computational cost may increase, they are generally more accurate than the separated methods. In addition to improving accuracy, recent researchers have also focused on enhancing the robustness of algorithms across various applications \cite{schmidt2003robust, furrer2018evaluation, sarabandi2022hand, wu2019hand}. However, these efforts were mostly focused on scenarios characterized by comparatively high-precision pose data. This differs from our use case where the trajectories are substantially affected by noise or accumulated error. In this paper, we utilize a dual quaternion scheme similar to \cite{daniilidis1999hand, furrer2018evaluation} and construct a robust linear solving system using multiple constraints from screw theory to address the challenges from the VO/VIO trajectory. 

\subsection{Temporal Alignment}
Given that the hand and the eye sensors usually operate on different clocks, the temporal correspondence between $\mathbf{A}$ and $\mathbf{B}$ is usually unknown. The time alignment prior to spatial calibration is thus necessary. Kelly and Sukhatme \cite{kelly2014general} considered the problem to be a registration task and solved it by utilizing the iterative closest point algorithm. Based on the discrete Fourier transform (DFT) theory, the time alignment of trajectories can be converted to the correlation analysis between two invariant signals derived from screw motion. This simple and effective method has been widely used in data synchronization for hand-eye calibration \cite{furrer2018evaluation, ackerman2013sensor, li2015simultaneous, pachtrachai2018chess}. However, the precision of this method is limited by the temporal resolution of the correlation function, with low-frequency data resulting in reduced time alignment accuracy. Alternatively, the TUM-VI benchmark \cite{schuberttum} achieves time alignment by utilizing information from an error function, which is calculated through a grid search between the motion invariants. Additionally, a parabolic fitting is applied to the error function to enhance the precision of the time alignment. Unlike the grid search method in the time domain employed in TUM-VI, our approach builds upon the more commonly used DFT-based approach. To overcome the limitation posed by data frequency, we adopt a similar technique inspired by TUM-VI. 

In some studies of continuous-time state estimation, by parameterizing the state variables as continuous-time functions, it is possible to achieve simultaneous spatiotemporal multi-sensor calibration within a MLE framework \cite{furgale2012continuous, rehder2016general, sommer2016continuous}. These high-precision methods can be easily extended to the pose-only scenario, but they are sensitive to the initial condition. In our work, we use the result of our linear calibration as the initial guess and perform a further refinement. 

\section{METHODOLOGY} \label{SEC: METHODOLOGY}

\begin{figure*}[ht]
    \centering
    \includegraphics[width=0.8\linewidth]{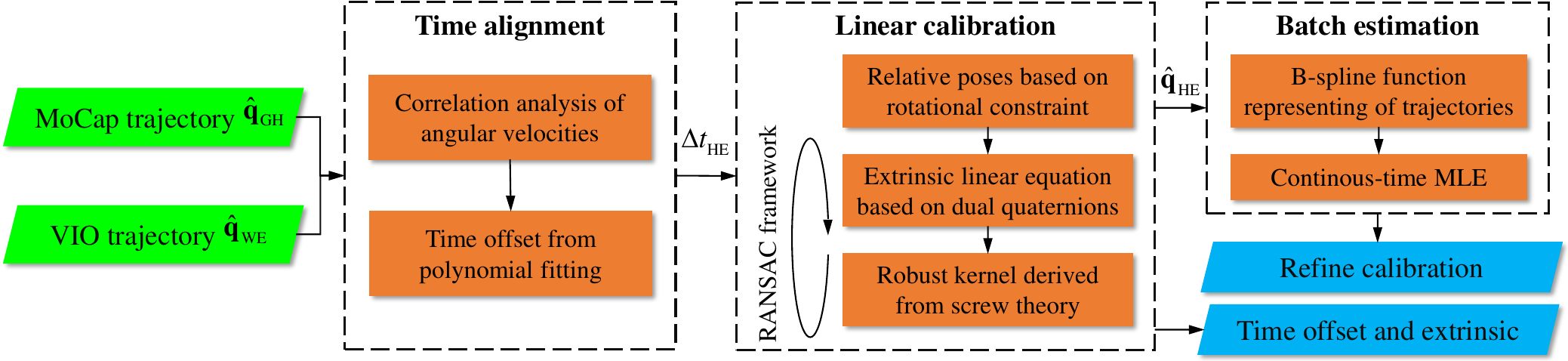}
    \caption{Flowchart of the proposed spatiotemporal hand-eye calibration, where the green and blue parallelograms represent the inputs and outputs respectively, and the orange rectangles represent the critical processing steps. }
    \label{FIG: Flowchart of the proposed algorithm}
\end{figure*}

To better illustrate our algorithm, we take our hand-eye calibration platform as an example (see in Fig. \ref{FIG: Our hand-eye calibration platform}). The unit dual quaternions $\hat{\mathbf{q}} \in \mathbb{DQ}$ is used to represent the homogeneous transformation. A dual quaternion $\hat{\mathbf{q}}$ has the  form $\hat{\mathbf{q}} = \mathbf{q} + \epsilon\mathbf{q}^{\prime} = \left(\mathbf{q}, \mathbf{q}^{\prime}\right)$, where Hamiltonian quaternions $\mathbf{q}$ and $\mathbf{q}^{\prime}$ are the standard part and the dual part of $\hat{\mathbf{q}}$ respectively, and $\epsilon$ is the infinitesimal unit satisfying $\epsilon^2 = 0$. Our goal is to estimate the time offset $\Delta t_{\scriptscriptstyle\mathrm{HE}}$ and the extrinsic $\hat{\mathbf{q}}_{\scriptscriptstyle\mathrm{HE}}$ between the MoCap (hand) trajectory $\hat{\mathbf{q}}_{\scriptscriptstyle\mathrm{GH}}$ and the estimated VIO (eye) trajectory $\hat{\mathbf{q}}_{\scriptscriptstyle\mathrm{WE}}$. Fig. \ref{FIG: Flowchart of the proposed algorithm} provides an overview of the proposed algorithm, which comprises three modules. The main steps of the algorithm will be described in the remaining of this section.

\subsection{Time Alignment} \label{SUBSEC: Time Alignment}
\begin{figure}[t]
    \centering
    \includegraphics[width=0.95\linewidth]{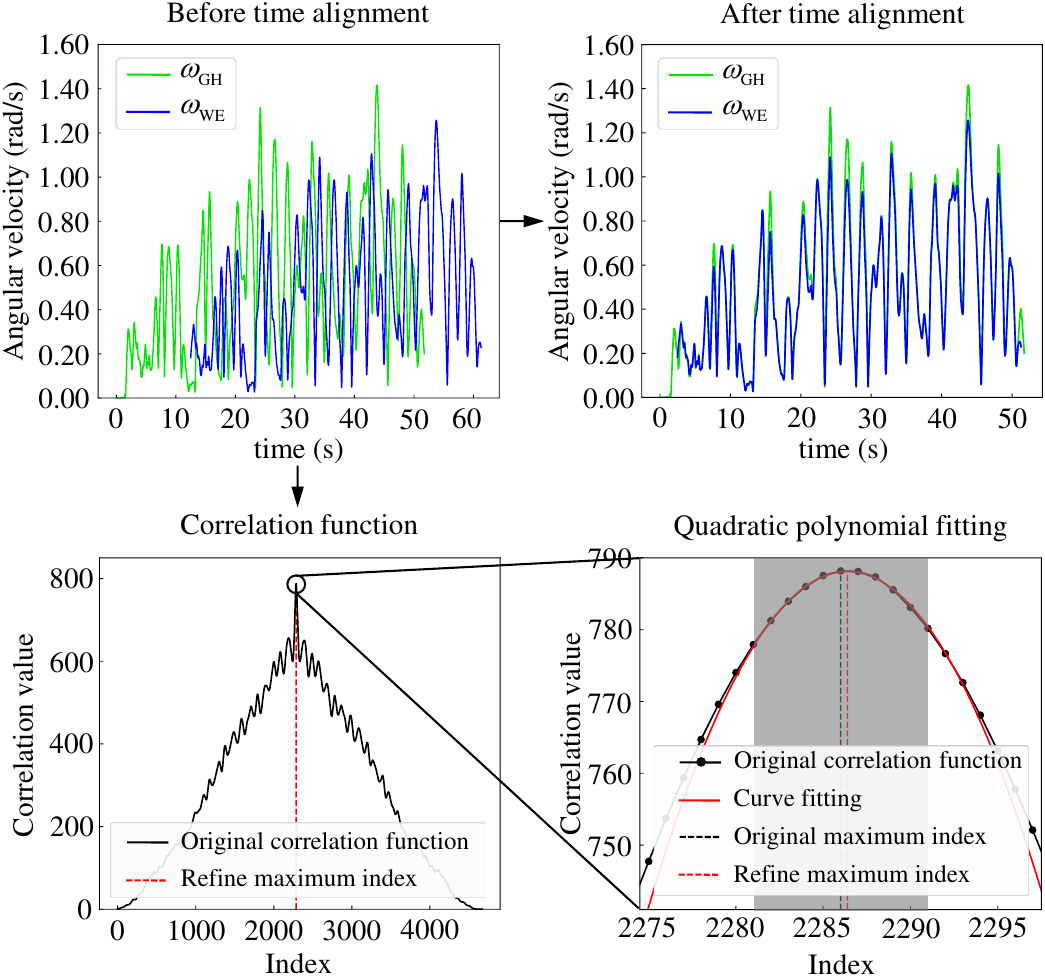}
    \caption{Illustration of time alignment. The time offset can be determined by performing a quadratic polynomial curve fitting around the maximum (highlighted in gray) of the correlation function and obtaining the index of the maximum. This method enables synchronization of angular velocity at a finer granularity. }
    \label{FIG: Instruction of time alignment}
\end{figure}

In order to process poses from sensors with different clocks, the first crucial step is to align the two sets of timestamps. To achieve this, we can leverage the constraint from screw motion, i.e., based on the equality of the angular velocities $\omega_{\scriptscriptstyle\mathrm{GH}}$ and $\omega_{\scriptscriptstyle\mathrm{WE}}$ of trajectories $\hat{\mathbf{q}}_{\scriptscriptstyle\mathrm{GH}}$ and $\hat{\mathbf{q}}_{\scriptscriptstyle\mathrm{WE}}$, which is independent of calibration parameters. This simplifies the time alignment to the synchronization of angular velocity signals. Based on theory of DFT, the correlation between two time domain signals is greater when they exhibit higher similarity. Therefore, the synchronization involves finding the time shift, $\tau_{\mathrm{shift}}$, where the correlation function reaches its maximum:  
\begin{equation}
    \tau_{\mathrm{shift}} = \underset{\mathrm{index}}{\operatorname{argmax}} \left(\operatorname{Corr}\left(\omega_{\scriptscriptstyle\mathrm{GH}}, \omega_{\scriptscriptstyle\mathrm{WE}}\right)\right),
    \label{EQ: Time shift}
\end{equation}
where $\operatorname{Corr}\left(\cdot\right)$ is the correlation function. 

Given the discrete character of the angular velocity signals, the correlation function also appears discrete in the time domain, with the precision of the obtained time shift depending on the temporal resolution of the function. Assuming the data around the maximum follows a quadratic polynomial distribution, we perform curve fitting to refine our time shift similar to \cite{schuberttum}. As illustrated in Fig. \ref{FIG: Instruction of time alignment}, this approach allows us to accurately determine the index of maximum correlation and trace it back to the corresponding time offset, $\Delta t_{\scriptscriptstyle\mathrm{HE}}$. 

\subsection{Linear Calibration} \label{SUBSEC: Linear Calibration}
Given the time aligned trajectories $\hat{\mathbf{q}}_{\scriptscriptstyle\mathrm{GH}}$ and $\hat{\mathbf{q}}_{\scriptscriptstyle\mathrm{WE}}$, we can perform spatial hand-eye calibration. As shown in Fig. \ref{FIG: Our hand-eye calibration platform}, for any hand-eye motion from $i$ to $j$ in the trajectories, we have: 
\begin{equation}
    \hat{\mathbf{q}}_{\scriptscriptstyle\mathrm{GH}_i} \hat{\mathbf{q}}_{\scriptscriptstyle\mathrm{HE}} \hat{\mathbf{q}}_{\scriptscriptstyle\mathrm{WE}_i}^{-1} = \hat{\mathbf{q}}_{\scriptscriptstyle\mathrm{GH}_j} \hat{\mathbf{q}}_{\scriptscriptstyle\mathrm{HE}} \hat{\mathbf{q}}_{\scriptscriptstyle\mathrm{WE}_j}^{-1}. 
    \label{EQ: Hand-eye motion}
\end{equation}

Using the relative transformation between two poses, i.e., $\hat{\mathbf{q}}_{\scriptscriptstyle\mathrm{H}_i\scriptscriptstyle\mathrm{H}_j} = \hat{\mathbf{q}}_{\scriptscriptstyle\mathrm{GH}_i}^{-1} \hat{\mathbf{q}}_{\scriptscriptstyle\mathrm{GH}_j}$ and $\hat{\mathbf{q}}_{\scriptscriptstyle\mathrm{E}_i\scriptscriptstyle\mathrm{E}_j} = \hat{\mathbf{q}}_{\scriptscriptstyle\mathrm{WE}_i}^{-1} \hat{\mathbf{q}}_{\scriptscriptstyle\mathrm{WE}_j}$, \eqref{EQ: Hand-eye motion} can be rewritten in the form of $\mathbf{AX}$ = $\mathbf{XB}$ as $\hat{\mathbf{q}}_{\scriptscriptstyle\mathrm{H}_i\scriptscriptstyle\mathrm{H}_j} \hat{\mathbf{q}}_{\scriptscriptstyle\mathrm{HE}} = \hat{\mathbf{q}}_{\scriptscriptstyle\mathrm{HE}} \hat{\mathbf{q}}_{\scriptscriptstyle\mathrm{E}_i\scriptscriptstyle\mathrm{E}_j}$. This fundamental equation can be divided into the standard and dual parts, yielding: 
\begin{equation}
\begin{aligned}
    {\mathbf{q}}_{\scriptscriptstyle\mathrm{H}_i\scriptscriptstyle\mathrm{H}_j} {\mathbf{q}}_{\scriptscriptstyle\mathrm{HE}} - {\mathbf{q}}_{\scriptscriptstyle\mathrm{HE}} {\mathbf{q}}_{\scriptscriptstyle\mathrm{E}_i\scriptscriptstyle\mathrm{E}_j} &= 0, \\ 
    {\mathbf{q}}_{\scriptscriptstyle\mathrm{H}_i\scriptscriptstyle\mathrm{H}_j}^{\prime} {\mathbf{q}}_{\scriptscriptstyle\mathrm{HE}} - {\mathbf{q}}_{\scriptscriptstyle\mathrm{HE}} {\mathbf{q}}_{\scriptscriptstyle\mathrm{E}_i\scriptscriptstyle\mathrm{E}_j}^{\prime} + {\mathbf{q}}_{\scriptscriptstyle\mathrm{H}_i\scriptscriptstyle\mathrm{H}_j} {\mathbf{q}}_{\scriptscriptstyle\mathrm{HE}}^{\prime} - {\mathbf{q}}_{\scriptscriptstyle\mathrm{HE}}^{\prime} {\mathbf{q}}_{\scriptscriptstyle\mathrm{E}_i\scriptscriptstyle\mathrm{E}_j} &= 0. 
    \label{EQ: Standard and dual parts of fundamental equation}
\end{aligned}
\end{equation}

Due to the redundancy of the scalar part of the dual quaternion, we set $\mathbf{r} = \bigl({\mathbf{q}}_{\scriptscriptstyle\mathrm{H}_i\scriptscriptstyle\mathrm{H}_j}\bigr)_v$, $\mathbf{r}^{\prime} = \bigl({\mathbf{q}}_{\scriptscriptstyle\mathrm{H}_i\scriptscriptstyle\mathrm{H}_j}^{\prime}\bigr)_v$, $\mathbf{s} = \bigl({\mathbf{q}}_{\scriptscriptstyle\mathrm{E}_i\scriptscriptstyle\mathrm{E}_j}\bigr)_v$ and $\mathbf{s}^{\prime} = \bigl({\mathbf{q}}_{\scriptscriptstyle\mathrm{E}_i\scriptscriptstyle\mathrm{E}_j}^{\prime}\bigr)_v$, where $\left(\cdot\right)_v$ denotes the vector part of the quaternion. The linear equation for extrinsic calibration derived from a single motion can be written as: 
\begin{equation}
    \begin{bmatrix}
    \mathbf{r}-\mathbf{s}  &\left(\mathbf{r}+\mathbf{s}\right)^\wedge  &\mathbf{0}_{3\times1}   &\mathbf{0}_{3\times3} \\
    \mathbf{r}^{\prime}-\mathbf{s}^{\prime}  &\left(\mathbf{r}^{\prime}+\mathbf{s}^{\prime}\right)^\wedge  &\mathbf{r}-\mathbf{s}  &\left(\mathbf{r}+\mathbf{s}\right)^\wedge
    \end{bmatrix}
    \begin{bmatrix}
    {\mathbf{q}}_{\scriptscriptstyle\mathrm{HE}} \\
    {\mathbf{q}}_{\scriptscriptstyle\mathrm{HE}}^{\prime}
    \end{bmatrix}
    =0, 
    \label{EQ: Linear equation for extrinsic calibration}
\end{equation}
where $\left(\cdot\right)^\wedge$ denotes the antisymmetric matrix of a vector, and the coefficient matrix with dimensions of $6\times8$ will be denoted as $\mathbf{S}$. With $n\ge2$ motions, we can stack $\mathbf{S}$ to obtain a $6n\times8$ matrix with rank 6 in the noise-free case as: 
\begin{equation}
    \mathbf{M} = \left[\mathbf{S}_1^\mathrm{T}, \mathbf{S}_2^\mathrm{T}, \ldots,\mathbf{S}_n^\mathrm{T}\right]^\mathrm{T}. 
    \label{EQ: Linear calibration matrix}
\end{equation}

The singular value decomposition (SVD) algorithm is then used to find the linear least squares solution of the extrinsic with the constraint of the unit dual quaternion. For more detailed information about the fundamental principles of dual quaternion-based hand-eye calibration, please refer to \cite{daniilidis1999hand}.

SVD is known to be sensitive to noise and outliers. In the following, we will propose three strategies to enhance the accuracy and robustness of the algorithm.

\subsubsection{Relative poses construction} As shown in \eqref{EQ: Linear equation for extrinsic calibration}, the equation entirely relies on the relative poses of the hand-eye motion, therefore, the relative poses construction method will significantly affect the quality of the solution. Conventional methods either use a global or an inter-frame strategy. The former fixes a certain frame and calculates the relative poses of the remaining frames with respect to it. However, this method is prone to coupling the trajectory drift in the VO/VIO scenario. The latter calculates the relative poses between two successive frames, but may suffer from the noise caused by insufficient motion. Moreover, for a relative pose with pure translation, the matrix in \eqref{EQ: Linear equation for extrinsic calibration} will degenerate and cannot constrain the dual part of the extrinsic.

\begin{figure}[t]
    \centering
    \includegraphics[width=0.68\linewidth]{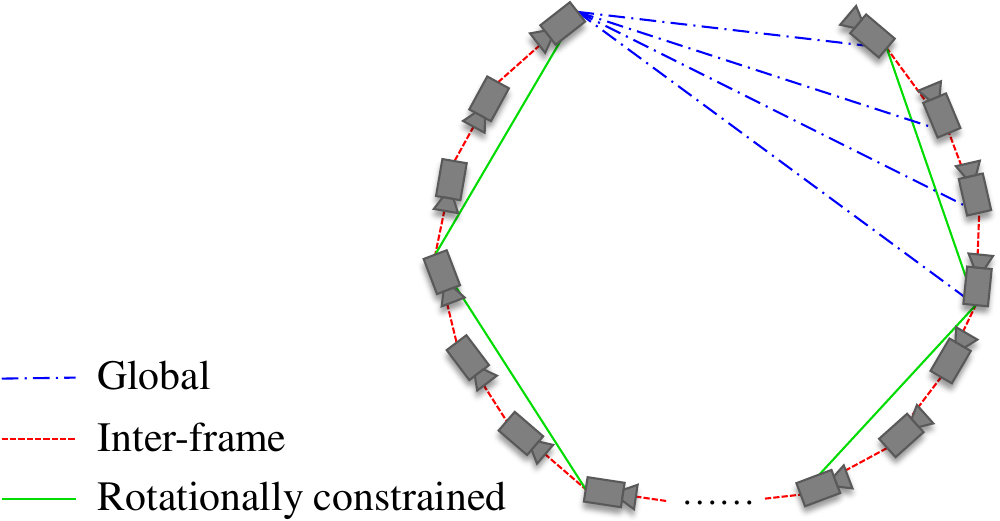}
    \caption{Illustration of relative poses construction, different methods are represented by three different lines, and ours shown as the solid green line. }
    \label{Fig: Relative pose construction}
\end{figure}

We use a rotationally constrained approach to select the hand-eye keyframes to construct the relative poses, aiming to mitigate the error coupling and solution degeneracy. Specifically, one can keep searching forward from a frame $\hat{\mathbf{q}}_i$ in the hand or eye trajectory, and build the relative pose when a frame $\hat{\mathbf{q}}_j$ satisfies the constraint: 
\begin{equation}
    2\arccos\left(\left(\hat{\mathbf{q}}_i^{-1}\hat{\mathbf{q}}_j\right)_w\right) \ge \eta, 
    \label{EQ: rotational constraint}
\end{equation}
where $\left(\cdot\right)_w$ denotes the scalar part of the standard component of the dual quaternion, and $\eta$ is an adjustable threshold which we set to 5 degrees. Fig. \ref{Fig: Relative pose construction} illustrates the construction of relative poses using both conventional methods and our proposed method for comparison. 

\subsubsection{Robust kernel} Despite the effort to construct high-quality relative poses, they still contain varying degrees of noise. To quantify the weights of different relative poses, we propose a robust kernel in the linear system construction. 

For a unit dual quaternion $\hat{\mathbf{q}}$, its scalar part is defined as: 
\begin{equation}
    \operatorname{Scalar}\left(\hat{\mathbf{q}}\right) = \frac{\left(\hat{\mathbf{q}} + \hat{\mathbf{q}}^{-1}\right)}{2}, 
    \label{EQ: Scalar part definition}
\end{equation}
which can be expressed in the form of a vector, written as: 
\begin{equation}
\begin{aligned}
    \operatorname{Scalar}\left(\hat{\mathbf{q}}\right) &= \left[\omega, \mathbf{0}_{1\times3}, \omega^\prime, \mathbf{0}_{1\times3}\right]^\mathrm{T} \\ 
    &= \Bigl[\cos\frac{\theta}{2}, \mathbf{0}_{1\times3}, -\frac{d}{2}\sin\frac{\theta}{2}, \mathbf{0}_{1\times3}\Bigr]^\mathrm{T}, 
    \label{EQ: Scalar part in vector form}
\end{aligned}
\end{equation}
where $\omega$ and $\omega^\prime$ are the scalar parts of the standard part and the dual part of the dual quaternion respectively, $\theta$ denotes the rotation angle, and $d$ represents the translation norm of the screw motion. 

The screw concatenation between the hand and eye can be further transformed from the form of $\mathbf{AX}$ = $\mathbf{XB}$ to $\hat{\mathbf{q}}_{\scriptscriptstyle\mathrm{H}_i\scriptscriptstyle\mathrm{H}_j} = \hat{\mathbf{q}}_{\scriptscriptstyle\mathrm{HE}} \hat{\mathbf{q}}_{\scriptscriptstyle\mathrm{E}_i\scriptscriptstyle\mathrm{E}_j} \hat{\mathbf{q}}_{\scriptscriptstyle\mathrm{HE}}^{-1}$. Based on the definition of the scalar part in \eqref{EQ: Scalar part definition}, we have: 
\begin{equation}
\begin{aligned}
    \operatorname{Scalar} \left(\hat{\mathbf{q}}_{\scriptscriptstyle\mathrm{H}_i\scriptscriptstyle\mathrm{H}_j}\right) &= \frac{1}{2} \left(\hat{\mathbf{q}}_{\scriptscriptstyle\mathrm{H}_i\scriptscriptstyle\mathrm{H}_j} + \hat{\mathbf{q}}_{\scriptscriptstyle\mathrm{H}_i\scriptscriptstyle\mathrm{H}_j}^{-1}\right) \\ 
    &= \frac{1}{2} \left(\hat{\mathbf{q}}_{\scriptscriptstyle\mathrm{HE}} \hat{\mathbf{q}}_{\scriptscriptstyle\mathrm{E}_i\scriptscriptstyle\mathrm{E}_j} \hat{\mathbf{q}}_{\scriptscriptstyle\mathrm{HE}}^{-1} + \hat{\mathbf{q}}_{\scriptscriptstyle\mathrm{HE}} \hat{\mathbf{q}}_{\scriptscriptstyle\mathrm{E}_i\scriptscriptstyle\mathrm{E}_j}^{-1} \hat{\mathbf{q}}_{\scriptscriptstyle\mathrm{HE}}^{-1}\right) \\ 
    &= \frac{1}{2} \hat{\mathbf{q}}_{\scriptscriptstyle\mathrm{HE}} \left(\hat{\mathbf{q}}_{\scriptscriptstyle\mathrm{E}_i\scriptscriptstyle\mathrm{E}_j} + \hat{\mathbf{q}}_{\scriptscriptstyle\mathrm{E}_i\scriptscriptstyle\mathrm{E}_j}^{-1}\right) \hat{\mathbf{q}}_{\scriptscriptstyle\mathrm{HE}}^{-1} \\ 
    &= \hat{\mathbf{q}}_{\scriptscriptstyle\mathrm{HE}} \operatorname{Scalar} \left(\hat{\mathbf{q}}_{\scriptscriptstyle\mathrm{E}_i\scriptscriptstyle\mathrm{E}_j}\right) \hat{\mathbf{q}}_{\scriptscriptstyle\mathrm{HE}}^{-1} \\ 
    &=\operatorname{Scalar} \left(\hat{\mathbf{q}}_{\scriptscriptstyle\mathrm{E}_i\scriptscriptstyle\mathrm{E}_j}\right) \hat{\mathbf{q}}_{\scriptscriptstyle\mathrm{HE}} \hat{\mathbf{q}}_{\scriptscriptstyle\mathrm{HE}}^{-1} \\
    &= \operatorname{Scalar} \left(\hat{\mathbf{q}}_{\scriptscriptstyle\mathrm{E}_i\scriptscriptstyle\mathrm{E}_j}\right), 
    \label{EQ: Screw congruence theorem}
\end{aligned}
\end{equation}
which demonstrates that the scalar parts of the hand-eye relative pose pair are completely equal in the absence of noise. According to \eqref{EQ: Scalar part in vector form}, this constraint can be transformed into the equality of rotation angles in local frames and translation norms along the principal axes of rotation, known as the screw congruence theorem. 

We design the function as \eqref{EQ: Evaluation function} to evaluate the quality of the hand-eye relative poses based on the scalar part of the dual quaternion in \eqref{EQ: Scalar part in vector form}. The function yields a result of 1 when the screw motion constraint in \eqref{EQ: Screw congruence theorem} is strictly satisfied, while deviating from 1 as the noise increases. 
\begin{equation}
\begin{aligned}
    &E_i\left(\hat{\mathbf{q}}_{\scriptscriptstyle\mathrm{H}_i\scriptscriptstyle\mathrm{H}_j}, \hat{\mathbf{q}}_{\scriptscriptstyle\mathrm{E}_i\scriptscriptstyle\mathrm{E}_j}\right) \\
    = &E_i^\prime \left(\operatorname{Scalar} \left(\hat{\mathbf{q}}_{\scriptscriptstyle\mathrm{H}_i\scriptscriptstyle\mathrm{H}_j}\right), \operatorname{Scalar} \left(\hat{\mathbf{q}}_{\scriptscriptstyle\mathrm{E}_i\scriptscriptstyle\mathrm{E}_j}\right)\right) \\
    = &\frac{1}{2} \left(\frac{\max\left(\left|\omega_{\scriptscriptstyle\mathrm{H}_i\scriptscriptstyle\mathrm{H}_j}\right|, \left|\omega_{\scriptscriptstyle\mathrm{E}_i\scriptscriptstyle\mathrm{E}_j}\right|\right)}{\min\left(\left|\omega_{\scriptscriptstyle\mathrm{H}_i\scriptscriptstyle\mathrm{H}_j}\right|, \left|\omega_{\scriptscriptstyle\mathrm{E}_i\scriptscriptstyle\mathrm{E}_j}\right|\right)} + \frac{\max\bigl(\bigl|\omega_{\scriptscriptstyle\mathrm{H}_i\scriptscriptstyle\mathrm{H}_j}^{\prime}\bigr|, \bigl|\omega_{\scriptscriptstyle\mathrm{E}_i\scriptscriptstyle\mathrm{E}_j}^{\prime}\bigr|\bigr)}{\min\bigl(\bigl|\omega_{\scriptscriptstyle\mathrm{H}_i\scriptscriptstyle\mathrm{H}_j}^{\prime}\bigr|, \bigl|\omega_{\scriptscriptstyle\mathrm{E}_i\scriptscriptstyle\mathrm{E}_j}^{\prime}\bigr|\bigr)}\right). 
    \label{EQ: Evaluation function}
\end{aligned}
\end{equation}

Based on \eqref{EQ: Evaluation function}, we can define the robust kernel $W$ as: 
\begin{equation}
    W_i\left(\hat{\mathbf{q}}_{\scriptscriptstyle\mathrm{H}_i\scriptscriptstyle\mathrm{H}_j}, \hat{\mathbf{q}}_{\scriptscriptstyle\mathrm{E}_i\scriptscriptstyle\mathrm{E}_j}\right) = \exp\left(\mu\left(1 - E_i\left(\hat{\mathbf{q}}_{\scriptscriptstyle\mathrm{H}_i\scriptscriptstyle\mathrm{H}_j}, \hat{\mathbf{q}}_{\scriptscriptstyle\mathrm{E}_i\scriptscriptstyle\mathrm{E}_j}\right)^2\right)\right),  
    \label{EQ: Robust kernel}
\end{equation}
where the parameter $\mu$ is an adjustable magnification factor, which we set to 5. Hence, the linear calibration matrix in \eqref{EQ: Linear calibration matrix} can be robustified for better numerical stability: 
\begin{equation}
    \mathbf{M}_{\mathrm{r}} = \left[W_1\mathbf{S}_1^\mathrm{T}, W_2\mathbf{S}_2^\mathrm{T}, \ldots, W_n\mathbf{S}_n^\mathrm{T}\right]^\mathrm{T}. 
    \label{EQ: Robust linear calibration matrix}
\end{equation}

\subsubsection{Outlier elimination} During the calibration process, it is also important to identify and discard outliers that exhibit significant error. We incorporate the RANSAC algorithm which utilizes an iterative sampling strategy, and has the ability to recover the inliers from noisy data. 

Note that at least two pairs of relative poses are required to construct the linear calibration system. For each iteration, we randomly sample a pose pair subset consisting of two members and determine inliers by using the quantitative criterion as: 
\begin{equation}
\begin{aligned}
    e_1 &= \left(\hat{\mathbf{q}}_{\scriptscriptstyle\mathrm{HE}} \hat{\mathbf{q}}_{\scriptscriptstyle\mathrm{E}_i\scriptscriptstyle\mathrm{E}_j} \hat{\mathbf{q}}_{\scriptscriptstyle\mathrm{HE}}^{-1} \hat{\mathbf{q}}_{\scriptscriptstyle\mathrm{H}_i\scriptscriptstyle\mathrm{H}_j}^{-1}\right)_{\operatorname{rot}} < \varphi, \\
    e_2 &= \left(\hat{\mathbf{q}}_{\scriptscriptstyle\mathrm{HE}} \hat{\mathbf{q}}_{\scriptscriptstyle\mathrm{E}_i\scriptscriptstyle\mathrm{E}_j} \hat{\mathbf{q}}_{\scriptscriptstyle\mathrm{HE}}^{-1} \hat{\mathbf{q}}_{\scriptscriptstyle\mathrm{H}_i\scriptscriptstyle\mathrm{H}_j}^{-1}\right)_{\operatorname{trans}}
    \label{EQ: Inlier quantitative criterion} < \psi, 
\end{aligned}
\end{equation}
where $\left(\cdot\right)_{\operatorname{rot}}$ and $\left(\cdot\right)_{\operatorname{trans}}$ are the rotation angle and translation norm of the dual quaternion, respectively. Inliers are required to have both indicators in \eqref{EQ: Inlier quantitative criterion} less than the thresholds $\varphi$ and $\psi$, which we set to 0.5 degrees and 0.02 m experimentally. 

\begin{algorithm}[t]
    \SetAlgoLined
    \KwIn{Time aligned hand-eye trajectories $\hat{\mathbf{q}}_{\scriptscriptstyle\mathrm{GH}}$, $\hat{\mathbf{q}}_{\scriptscriptstyle\mathrm{WE}}$.}
    \KwOut{Hand-eye extrinsic $\hat{\mathbf{q}}_{\scriptscriptstyle\mathrm{HE}}^{\ast}$.}
    $\left\{\hat{\mathbf{q}}_{\scriptscriptstyle\mathrm{E}_i\scriptscriptstyle\mathrm{E}_j}, \hat{\mathbf{q}}_{\scriptscriptstyle\mathrm{H}_i\scriptscriptstyle\mathrm{H}_j}\right\} \in \mathcal{C} \gets$ Construct the relative poses based on the rotational constraint in \eqref{EQ: rotational constraint}\;
    \While{not reached the iteration limit}{
        Subset $\mathcal{D} \gets$ Randomly sampling from $\mathcal{C}$\;
        Construct $\mathbf{M}_{12\times8}$ based on $\mathcal{D}$ using \eqref{EQ: Linear calibration matrix}\;
        $\hat{\mathbf{q}}_{\scriptscriptstyle\mathrm{HE}}^{\mathrm{init}} \gets$ Apply SVD to M\;
        \ForEach{pose pair $\in$ $\mathcal{C}$}{
            $\mathcal{G} \gets$ Select inliers using \eqref{EQ: Inlier quantitative criterion} with $\hat{\mathbf{q}}_{\scriptscriptstyle\mathrm{HE}}^{\mathrm{init}}$\;
        }
        Construct $\mathbf{M}_{2n\times8}^{\prime}$ with robust kernel based on $\mathcal{G}$ using \eqref{EQ: Robust linear calibration matrix}\;
        $\hat{\mathbf{q}}_{\scriptscriptstyle\mathrm{HE}}^{\mathrm{refine}} \gets$ Apply SVD to $\mathbf{M}^{\prime}$\;
        $\hat{\mathbf{q}}_{\scriptscriptstyle\mathrm{HE}}^{\ast} \gets \hat{\mathbf{q}}_{\scriptscriptstyle\mathrm{HE}}^{\mathrm{refine}}$ with the smallest $\frac{\sigma_7}{\sigma_6}$ value;
    }
    \caption{Robust linear hand-eye calibration}
    \label{ALG: Linear hand-eye calibration}
\end{algorithm}

Additionally, we need to assess the quality of the solutions during the iteration. Given that the matrix $\mathbf{M}_{\mathrm{r}}$ in \eqref{EQ: Robust linear calibration matrix} possesses a two-dimensional null space, the singular values, $\sigma_7$ and $\sigma_8$, which are the last two in the descending diagonal matrix $\Sigma$ from the SVD result $\mathbf{M}_{\mathrm{r}} = \mathbf{U\Sigma V}^\mathrm{T}$, are expected to be zero in the absence of noise. However, when noise is present, these two singular values, corresponding to the noise, disproportionately increase compared to the remaining singular values. The ratio of $\sigma_7$ to the third-to-last singular value, $\sigma_6$, can serve as an indicator of this disproportionality, and a lower ratio suggests a reduced impact of noise. Consequently, we can establish a direct metric for quantifying the quality of the solutions, rather than relying on traditional measures such as the number of inliers or the root mean square error (RMSE) of the solutions with respect to inliers. The specific implementation of the linear spatial hand-eye calibration within the RANSAC framework is detailed in Algorithm \ref{ALG: Linear hand-eye calibration}. 

\subsection{Batch Estimation}
Despite the fact that our linear calibration method achieves good accuracy and robustness in the experiments of Section \ref{SEC: EXPERIMENTAL RESULTS}, it can be further improved by introducing the correlation between the temporal and spatial calibration parameters. In this section we follow the continuous-time batch estimation methods proposed in \cite{rehder2016general, sommer2016continuous} and provide an estimator within the rigorous theoretical framework of MLE, to jointly optimize the time offset and the spatial transformation. Since the estimator is hard to converge globally, we used the results derived in Section \ref{SUBSEC: Time Alignment} and \ref{SUBSEC: Linear Calibration} as the initial guesses. 

Specifically, the original hand trajectory is parameterized using a B-spline functions $\mathbf{T}_\mathrm{GH}\left(t\right)$, with the translation part represented by a B-spline in three-dimensional vector space, and the rotation part parameterized by a B-spline on $SO\left(3\right)$. For the B-spline $\mathbf{q}\left(t\right)$ on $SO\left(3\right)$ with order $\xi$, knots $\left\{t_i | i \in \left\{1, 2, 3, \dots, N\right\}\right\}$, and satisfying $N \ge 2\xi$, the function can be defined in each subinterval $\left\{t \in \left[t_i, t_{i+1}\right) | \xi \le i \le N - \xi\right\}$ as: 
\begin{equation}
    \mathbf{q}\left(t\right) = 
    \mathbf{q}_{l\left(i\right)} \prod_{j=1}^{\xi-1} \mathbf{EXP} \biggl(\Bigl(\sum_{k=\eta}^{i}f_k \Bigr)\mathbf{LOG} \left(\mathbf{q}_{\eta-1}^{-1} \mathbf{q}_{\eta}\right)\biggl), 
    \label{EQ: SO3 B-spline}
\end{equation}
where $l\left(i\right) = i - \xi + 1$ and $\eta = l\left(i\right) + j$, $f$ and $\mathbf{q}$ represent the B-spline basis functions and the control vertices in unit quaternion form, respectively. $\mathbf{EXP}$ denotes the mapping from the Lie algebra to the Lie group, while $\mathbf{LOG}$ represents the inverse process. 

The parameters determined by our estimator include $\mathbf{T}_{\scriptscriptstyle\mathrm{GH}}\left(t\right)$, homogeneous transformation $\mathbf{T}_{\scriptscriptstyle\mathrm{HE}}$, and time offset $\Delta t_{\scriptscriptstyle\mathrm{HE}}$. With the observations of hand-eye trajectories, we minimize the negative log-likelihood function as: 
\begin{equation} 
\begin{aligned}
    g &= \sum_{h=1}^{H-1} \rho\left(\left\|d\left(\mathbf{T}_{\scriptscriptstyle\mathrm{GH}}\left(t_h\right), \mathbf{T}_{\scriptscriptstyle\mathrm{GH}}\left(t_{h+1}\right), \mathbf{T}_{\scriptscriptstyle\mathrm{GH}_h}, \mathbf{T}_{\scriptscriptstyle\mathrm{GH}_{h+1}}\right)\right\|_{\Sigma_{\scriptscriptstyle\mathrm{H}}}^2\right) \\
    &+ \sum_{e=1}^{E-1} \rho\left(\left\|d\left(\mathbf{T}_{\scriptscriptstyle\mathrm{WE}}\left(t_e\right), \mathbf{T}_{\scriptscriptstyle\mathrm{WE}}\left(t_{e+1}\right), \mathbf{T}_{\scriptscriptstyle\mathrm{WE}_e}, \mathbf{T}_{\scriptscriptstyle\mathrm{WE}_{e+1}}\right)\right\|_{\Sigma_{\scriptscriptstyle\mathrm{E}}}^2\right), 
\end{aligned}
\end{equation}
where $H$ and $E$ denote the number of pose observations of the hand and the eye respectively, and $\rho\left(\cdot\right)$ is the Huber loss function. Due to the rigid connection between the hand and the eye, we have $\mathbf{T}_{\scriptscriptstyle\mathrm{WE}}\left(t\right) = \mathbf{T}_{\scriptscriptstyle\mathrm{WG}} \mathbf{T}_{\scriptscriptstyle\mathrm{GH}}\left(t - \Delta t_{\scriptscriptstyle\mathrm{HE}}\right) \mathbf{T}_{\scriptscriptstyle\mathrm{HE}}$. Meanwhile, we define the residual function in a relative form to eliminate the influence of the unknown $\mathbf{T}_{\scriptscriptstyle\mathrm{WG}}$ as: 
\begin{equation}
    d \left(\mathbf{T}_i, \mathbf{T}_{i+1}, \mathbf{T}_{i}^\prime, \mathbf{T}_{i+1}^\prime\right) = \mathbf{LOG} \Bigl(\mathbf{T}_i^{-1} \mathbf{T}_{i+1} \left(\mathbf{T}_{i}^{\prime-1}\mathbf{T}_{i+1}^\prime\right)^{-1}\Bigr). 
    \label{EQ: Residual function}
\end{equation}

For the above cost function, the Levenberg–Marquardt algorithm is used to obtain the refined calibration result. 

\section{EXPERIMENTAL RESULTS} \label{SEC: EXPERIMENTAL RESULTS}

\subsection{Ablation Studies}
To validate the effectiveness of the key improvement strategies proposed in our methodology, we conduct comparative experiments using the simulated datasets. Specifically, to obtain the required data, we model a real motion trajectory using the B-spline. Subsequently, we extract two trajectories from the model at 100Hz and 20Hz to simulate MoCap and VO/VIO trajectories for hand-eye calibration. Throughout the process, we introduce frame-wise cumulative error into each eye pose to simulate the noise and drift typically present in the VO/VIO trajectory. The standard deviations for translation and rotation errors are divided into ten levels, ranging from 0 to 5 mm and 0 to 0.2 degrees, respectively.

\begin{figure}[t]
    \centering
    \includegraphics[width=0.85\linewidth]{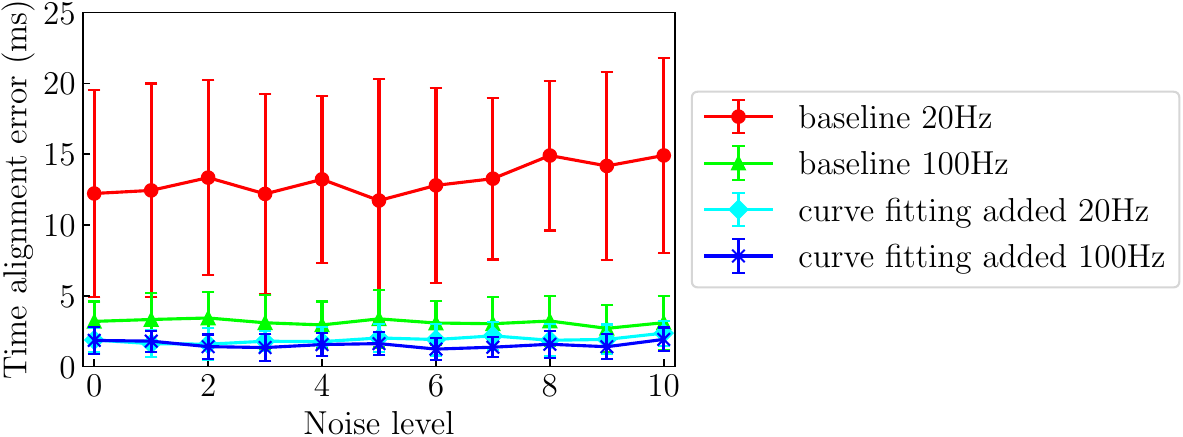}
    \caption{Performance comparison of different time alignment methods under various noise levels. We report the mean and standard deviation of the time alignment error. }
    \label{Fig: Performance of time alignment}
\end{figure}

We first test the time alignment algorithm proposed in Section \ref{SUBSEC: Time Alignment} to verify the enhancement provided by the quadratic polynomial curve fitting in the correlation analysis compared to the baseline approach used in \cite{furrer2018evaluation}. In this experiment, we introduce random delays into the eye trajectories derived from the B-spline to simulate the time offsets. To facilitate the correlation function computation, standardizing the frequencies of the trajectory pair is necessary. Specifically, we adjust the frequencies to match those of hand and eye trajectories, which are respectively 100Hz and 20Hz. As shown in Fig. \ref{Fig: Performance of time alignment}, correlation analysis-based time alignment methods exhibit minimal sensitivity to trajectory noise. However, limited by the temporal resolution of the correlation function, time alignment error of the baseline method is essentially determined by the trajectory frequency. Acquiring the high-frequency trajectory is challenging and requires significant computational effort. By implementing the curve fitting strategy, we can determine the maximum index of the correlation function with greater accuracy, leading to higher precision in time alignment, even at low frequency. 

\begin{figure}[t]
    \centering
    \includegraphics[width=\linewidth]{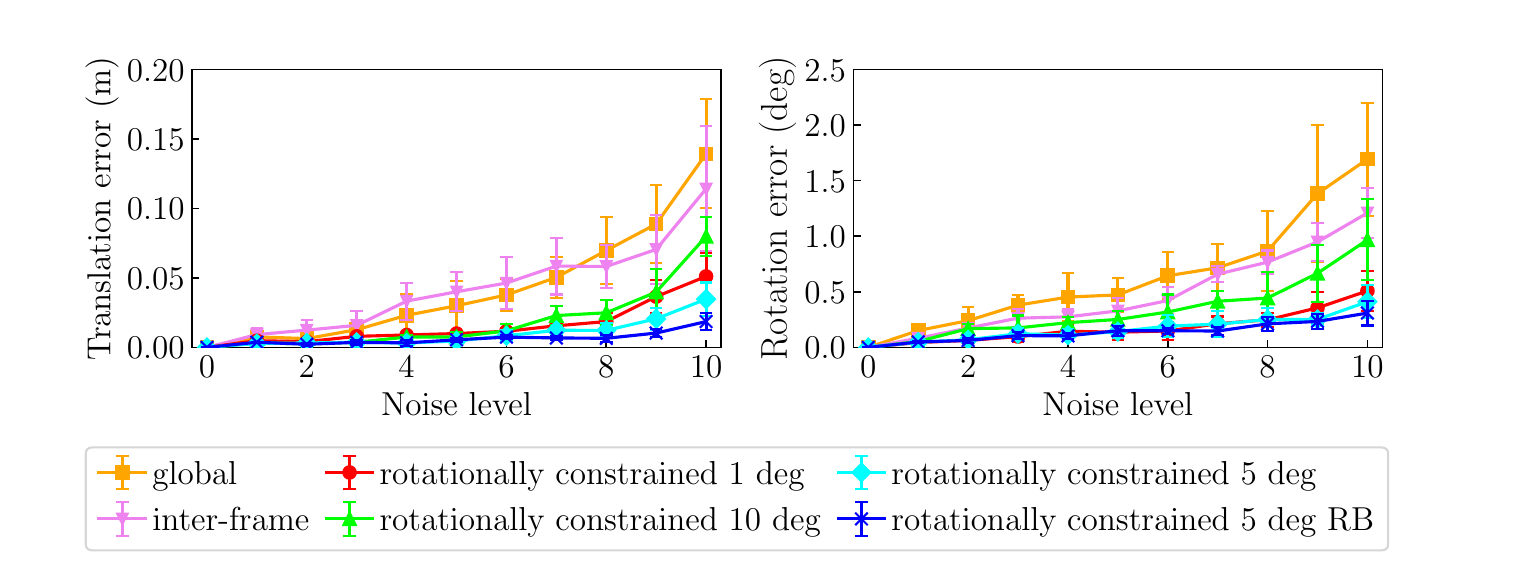}
    \caption{Performance comparison of different strategies used in linear calibration under various noise levels. We calculate the translation error and the rotation error separately, and report the mean and standard deviation. }
    \label{Fig: Performance of different strategies}
\end{figure}

Additionally, we evaluate the effectiveness of the strategies for spatial linear calibration proposed in Section \ref{SUBSEC: Linear Calibration}. The three relative poses construction methods, namely global, inter-frame, and rotationally constrained, are comparatively validated, and different thresholds in \eqref{EQ: rotational constraint} are explored within the proposed rotationally constrained approach. Meanwhile, we test the impact of the robust kernel (RB). In this experiment, the calibration error is used as a quantitative metric for assessment. Fig. \ref{Fig: Performance of different strategies} indicates that the calibration errors for all scenarios trend to increase as noise levels rise. Nonetheless, our method, which constructs rotationally constrained relative poses yields smaller error due to its ability to derive the linear equation with higher signal-to-noise ratio. At the same time, the rotational constraint threshold also affects the accuracy of the solution, with 5 degrees proving to be optimal in our tests. Furthermore, our robust kernel strategy demonstrates superior robustness in the presence of increasing noise, as indicated by the relatively small rise in calibration error and standard deviation. 

\subsection{Overall Evaluation}
We evaluate the performance of our hand-eye calibration algorithm using real-world datasets, including public VIO datasets, and datasets collected by our system (see in Fig. \ref{FIG: Our hand-eye calibration platform}). For comparison, we also evaluate two other state-of-the-art (SOTA) linear hand-eye calibration algorithms based on dual quaternion and random sampling, as implemented in \cite{furrer2018evaluation}, namely RANSAC scalar-based inlier check (RS) and RANSAC classic (RC). We use the well-known VIO algorithm OpenVINS \cite{geneva2020openvins} to estimate the trajectory of the eye, while the trajectory of the hand is derived from the ground-truth system. 

\begin{figure}[t]
    \centering
    \includegraphics[width=0.83\linewidth]{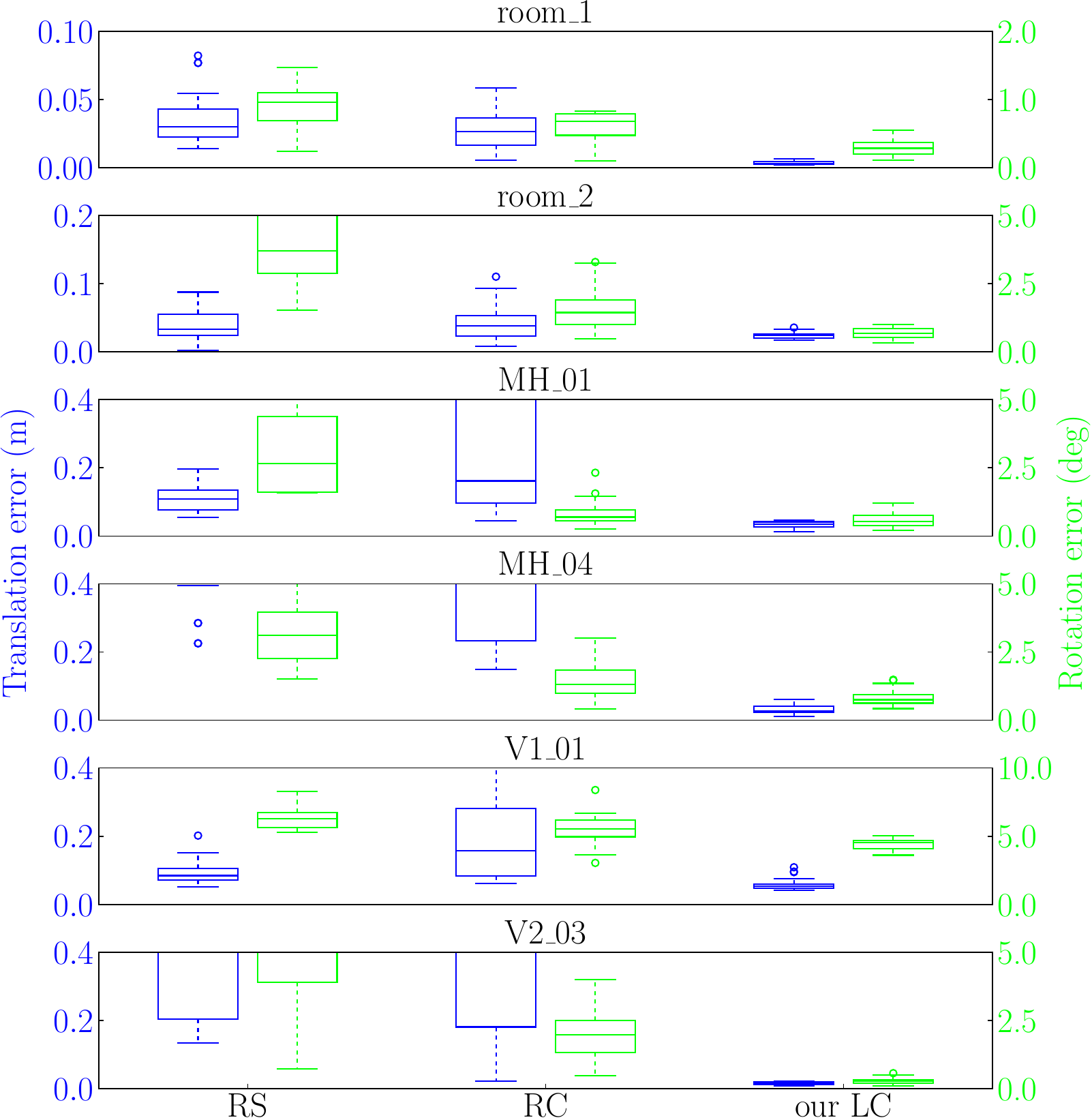}
    \caption{Performance comparison of different linear calibration methods on public datasets. We select two sequences in each scenario and count the distribution of the translation error and the rotation error separately. }
    \label{Fig: Public datasets boxplot}
\end{figure}

We first test on the TUM-VI room scenario \cite{schuberttum} as well as machine hall (MH) and Vicon room (V) scenarios provided by EuRoC benchmark \cite{burri2016euroc}. All three scenarios provide complete raw data of ground-truth trajectories, and achieve high-precision hand-eye calibration through the tightly-coupled method. We align the estimated VIO trajectories with the raw ground-truth trajectories using different hand-eye calibration algorithms. For evaluation metrics, we compare with the calibration results provided by the public datasets, and calculate the translation error and the rotation error. As shown in Fig. \ref{Fig: Public datasets boxplot}, we first evaluate our linear calibration (LC). The comparison results demonstrate that our algorithm achieves the highest accuracy and repeatability in all sequences. In challenging sequences such as MH\_04 and V2\_03, where the VIO algorithm struggles to deliver high-quality trajectories, traditional calibration algorithms are prone to fail, but our algorithm handles these situations effectively. 

\begin{table*}[t]
\caption{Detailed Performance of the Proposed Methods and the Trajectory Metrics on Public Datasets}
\label{Table: Detail Performance on public}
\begin{center}
\setlength{\tabcolsep}{5pt}
\renewcommand{\arraystretch}{0.9}
\begin{tabular}{c ccc ccc cc cc}
\toprule
\multirow{2}{*}[-0.2em]{Sequences} & \multicolumn{3}{c}{Error of our LC} & \multicolumn{3}{c}{Error of our BE} & \multicolumn{2}{c}{Original metrics} & \multicolumn{2}{c}{Our metrics} \\
\cmidrule(lr){2-4} \cmidrule(lr){5-7} \cmidrule(lr){8-9} \cmidrule(l){10-11}
                           & Time (ms) & Trans (m) & Rot (deg) & Time (ms) & Trans (m) & Rot (deg) & APE (m) & ARE (deg) & APE (m) & ARE (deg) \\
\midrule
Room\_1                    & 1.393     & 0.003     & 0.284     & 0.198     & 0.004     & 0.106     & 0.059   & 1.593     & 0.059   & 1.593     \\
Room\_6                    & 1.266     & 0.024     & 0.675     & 0.296     & 0.016     & 0.199     & 0.085   & 1.686     & 0.085   & 1.683     \\
MH\_01                     & 3.430     & 0.032     & 0.596     & 0.883     & 0.018     & 0.641     & 0.156   & 1.884     & 0.157   & 1.683     \\
MH\_04                     & 5.404     & 0.031     & 0.927     & 3.691     & 0.018     & 0.749     & 0.161   & 0.950     & 0.157   & 0.561     \\
V1\_01                     & 3.718     & 0.053     & 4.444     & 1.982     & 0.049     & 4.479     & 0.061   & 5.520     & 0.047   & 1.064     \\
V2\_03                     & 3.092     & 0.015     & 0.280     & 2.135     & 0.007     & 0.118     & 0.095   & 1.157     & 0.095   & 1.139     \\
\bottomrule
\end{tabular}
\end{center}
\end{table*}

\begin{figure}[t]
  \centering
  \subfloat[]{\includegraphics[width=0.8\linewidth]{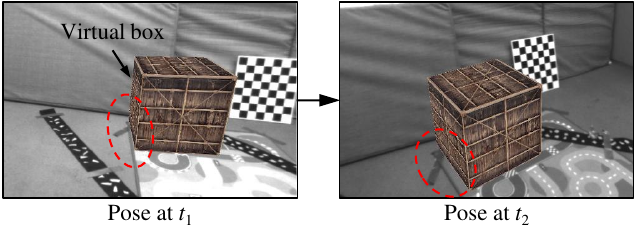}\label{fig: AR effects subfig 1}} \\
  \subfloat[]{\includegraphics[width=0.8\linewidth]{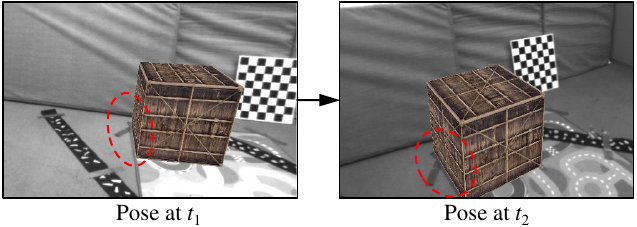}\label{fig: AR effects subfig 2}}
  \caption{AR visualization of the EuRoC V1\_01 sequence, featuring a virtual box rendered in the scene. This visualization utilizes (a) transformed ground-truth poses provided by the EuRoC benchmark and (b) transformed ground-truth poses based on the raw pre-calibrated trajectory and our calibration result. The red dashed circle highlights the difference in consistency between the virtual and real elements when using different poses. }
  \label{fig: Comparison of the AR effects}
\end{figure}

Table \ref{Table: Detail Performance on public} details the performance of our algorithm in each sequence, including the time alignment errors and the results of our batch estimation (BE). Additionally, to analyze the impact of hand-eye calibration on VIO trajectory evaluation, we use our calibration results to transform the raw ground-truth trajectories and calculate the absolute positional error (APE) and absolute rotational error (ARE) \cite{jinyu2019survey}, which are important metrics in VIO evaluation. As a comparison, we also calculate the original metrics using the transformed ground-truth provided by the public datasets. The obtained results demonstrate that our LC effectively accomplishes spatiotemporal hand-eye calibration, and our BE can further optimizes the result to achieve higher accuracy. In most of the sequences, our calibration algorithm provides accurate transformations for ground-truths with minimal impact on evaluation metrics. However, it is interesting to note that in some sequences, particularly V1\_01, our algorithm exhibits some errors. This may stem from the inherent inaccuracies in the calibration results of the benchmarks, corroborated by a similar conclusion in \cite{geneva2020openvins}. To provide a more intuitive illustration, we render a virtual box on the images of the V1\_01 sequence using two different transformed ground-truth trajectories. The first is provided by the EuRoC benchmark and the second is obtained based on our hand-eye calibration result. Inaccurate hand-eye calibration can lead to misalignment between the virtual and the real elements in this augmented reality (AR) application. As shown in the comparison results in Fig. \ref{fig: Comparison of the AR effects}, the virtual box is more consistent with the real world in our AR result, i.e., implying higher calibration accuracy. 

\begin{figure}[t]
    \centering
    \includegraphics[width=0.82\linewidth]{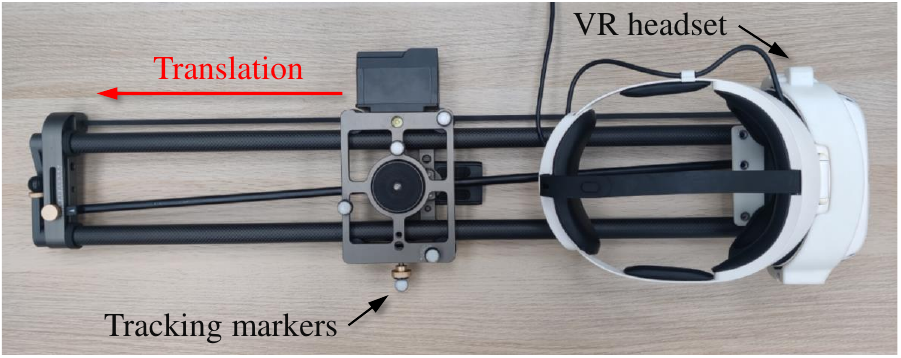}
    \caption{Experimental setup for evaluating the hand-eye calibration using a relative translation method. The VR headset remains stationary, and the tracking markers can be translated along the red arrow. }
    \label{Fig: Experimental setup. }
\end{figure}

\begin{table}[t]
\caption{Performance Comparison of Different Calibration Methods on Self-Collected Datasets. We Count the Translation Error at Different Relative Translation Distances}
\label{Table: Performance on self-Collected}
\centering
\begin{tabular}{
  @{\hspace{0.5em}}c!{}@{\hspace{1.0em}}c@{\hspace{1.0em}}c@{\hspace{1.0em}}c@{\hspace{1.0em}}c@{\hspace{0.5em}}
}
\toprule[1.0pt]
Sequences       & \begin{tabular}[c]{@{}c@{}}Error of \\ RS (m)\end{tabular} & \begin{tabular}[c]{@{}c@{}}Error of \\ RC (m)\end{tabular} & \begin{tabular}[c]{@{}c@{}}Error of \\ our\_LC (m)\end{tabular} & \begin{tabular}[c]{@{}c@{}}Error of \\ our\_BE (m)\end{tabular} \\
\midrule
0.1m\_easy      & 0.035                                                          & 0.028                                                          & 0.007                                                               & 0.003                                                               \\
0.2m\_easy      & 0.026                                                          & 0.022                                                          & 0.005                                                               & 0.003                                                               \\
0.3m\_easy      & 0.049                                                          & 0.034                                                          & 0.007                                                               & 0.002                                                               \\
0.1m\_difficult & 0.041                                                          & 0.045                                                          & 0.007                                                               & 0.004                                                               \\
0.2m\_difficult & 0.056                                                          & 0.038                                                          & 0.006                                                               & 0.004                                                               \\
0.3m\_difficult & 0.062                                                          & 0.047                                                          & 0.010                                                               & 0.004                                                               \\
\bottomrule[1.0pt]
\end{tabular}
\end{table}

In real-world system, evaluating hand-eye calibration is inherently difficult, as the ground-truth of offset is not available. To address this, we employ a relative approach using self-collected data. The experimental setup, as illustrated in Fig. \ref{Fig: Experimental setup. }, involved mounting a VR headset and tracking markers on the same object for motion. After obtaining a set of hand-eye trajectories, we translate the tracking markers in a specified direction to gather additional calibration data. We compute the norm of the relative translation using the extrinsics derived from the calibrations performed before and after the translation. Calibration error is then determined by comparing this norm against the high-precision, directly measured result.

Table \ref{Table: Performance on self-Collected} presents the errors obtained from different hand-eye calibration methods when the tracking markers are translated by 0.1 m, 0.2 m, and 0.3 m, respectively. We categorize the scenarios into easy (texture-rich and slow-moving) and difficult (texture-less and fast-moving) in the context of VIO, to test the robustness of the calibration algorithms. The experimental results show that our algorithm performs best in all sequences. Additionally, both our LC and BE achieve millimeter-level calibration accuracy and are less affected by trajectory error. 

\section{CONCLUSIONS} \label{SEC: CONCLUSIONS}

In this letter, we propose an improved spatiotemporal hand-eye calibration algorithm for trajectory alignment in VO/VIO evaluation. Aiming to optimize for VO/VIO scenarios, we have designed multiple strategies based on screw theory to enhance both the accuracy and robustness of our proposed algorithm. The validation experiments demonstrate that our algorithm outperforms SOTA methods, exhibiting superior accuracy while effectively mitigating the influence of noise. Our method is well poised to be applied in the evaluation of modern VO/VIO algorithms. Nevertheless, our algorithm is less effective over an extended period. This limitation arises partly because our calibration algorithm processes the entire trajectory, leading to inefficiency with long sequences. Furthermore, our algorithm presupposes a constant time offset, an assumption unsuitable over a long time. In future work, we will focus on spatiotemporal hand-eye calibration tailored for long-duration trajectories. The aim is to develop an algorithm that can process extensive data efficiently and address the problem of time offset drift over an extended period. 

\bibliographystyle{bib/IEEEtran}
\bibliography{bib/mybib}

\end{document}